\documentclass[11pt]{article}

\usepackage[preprint]{acl}

\usepackage{times}
\usepackage{latexsym}
\usepackage{tabularx}
\usepackage{enumitem}
\usepackage[most]{tcolorbox}
\usepackage{graphicx}

\usepackage[T1]{fontenc}

\usepackage[utf8]{inputenc}

\usepackage{microtype}

\usepackage{inconsolata}

\usepackage{graphicx}

\title{SLA-LENS: From Classroom Transcripts to CSLA-Informed and Evidence-Linked L2 Interaction Analytics}

\author{
  Nemika Tyagi$^{1}$\thanks{Equal contribution.} \thanks{Corresponding Author.}\quad
  Olga Kellert$^{1}$\footnotemark[1]\quad
  Phanos Kareen$^{1}$\footnotemark[1]\quad
  \textbf{Nelvin Licona-Guevara}$^1$\quad \\\\
  $^1$Arizona State University \quad \\
  \small{\texttt{\{ntyagi8, olga.kellert, kphanos, nliconag\}@asu.edu}}
}

\begin{document}
\maketitle

\begin{abstract}
What happens after a learner makes an error, and how does the instructor's response shape the next turn? Investigating these exchanges across classrooms requires context-sensitive annotation that is difficult to scale manually. We introduce SLA-LENS, an end-to-end LLM-assisted framework that connects classroom transcripts to structured pedagogical annotations and evidence-linked descriptive analytics. Grounded in interactionist Second Language Acquisition (SLA) theory, its annotation schema captures second language learner errors, pedagogical functions, and uptake or repair while accounting for speaker roles, conversational context, and hierarchical labels. We apply the framework to 40 authentic online L2 Spanish classroom transcripts comprising over 21,000 sentence-level instances. We evaluate proprietary and open-weight LLMs, including task-specific QLoRA adaptation, using expert-reviewed gold annotations, held-out silver data, and metrics sensitive to different types of annotation error. A deterministic reporting layer generates reproducible visualizations of participation, feedback, learner errors, uptake, and temporal interaction patterns, linked to supporting transcript evidence. By making these patterns accessible and inspectable, SLA-LENS supports systematic classroom discourse research, pedagogical reflection, and curriculum design
\footnote{Source code, documentation, aggregate corpus
statistics, and illustrative outputs are available at
\url{https://github.com/N3mika/SLA-LENS}.
The classroom transcripts and associated annotations
are restricted; research access may be considered
subject to institutional approval.}.

\end{abstract}

\section{Introduction}

What happens after a learner makes an error: correction, a hint, or a change of topic? Identifying these responses and the learner's subsequent uptake across classroom sessions is central to understanding the opportunities for learning that interaction provides. Interaction plays a central role in second-language acquisition (SLA)\cite{loewen2022}. In classroom settings, learner production, instructor feedback, and subsequent uptake or repair unfold across turns, creating opportunities for language development \citep{gass2015,long1996,schmidt1990,swain2005}. Understanding these exchanges is valuable for SLA research, pedagogical reflection, and the development of language-learning technologies. Online instruction makes such interaction increasingly observable. Platforms like Zoom preserve recordings, timestamps, speaker information, and automatically generated transcripts. Yet transcripts alone do not reveal what happened pedagogically \cite{loewen2022}. Systematically identifying pedagogical events across class transcripts requires context-sensitive discourse annotation \cite{demszky2023}. Learner errors may be identifiable within individual utterances, whereas feedback and uptake must be interpreted in relation to preceding turns. The following illustrative exchange demonstrates these dependencies:

\begin{quote}
\footnotesize
\setlength{\parskip}{2pt}

\noindent\textbf{Learner [\texttt{Er:G}]:}
\textit{Mi hermana tengo trece años.}\\
{\scriptsize English: My sister have(=error) thirteen years old.}

\noindent\textbf{Instructor [\texttt{F:EC}]:}
\textit{Con ``mi hermana'', usamos ``tiene'', no ``tengo''.}\\
{\scriptsize English: With ``my sister'', we use ``tiene'' (has), not ``tengo'' (have).}

\noindent\textbf{Instructor [\texttt{F:MF}]:}
\textit{El verbo debe concordar con el sujeto.}\\
{\scriptsize English: The verb must agree with the subject.}

\noindent\textbf{Learner [\texttt{R:Ack}]:}
\textit{Okay.}\\
{\scriptsize English: Okay.}

\noindent\textbf{Learner [\texttt{R:S}]:}
\textit{Mi hermana tiene trece años.}\\
{\scriptsize English: My sister is thirteen years old.}

\noindent\textbf{Instructor [\texttt{F:Pr}]:}
\textit{Muy bien.}\\
{\scriptsize English: Very good.}
\end{quote}

\begin{figure*}[t]
    \centering
    \includegraphics[width=\textwidth]{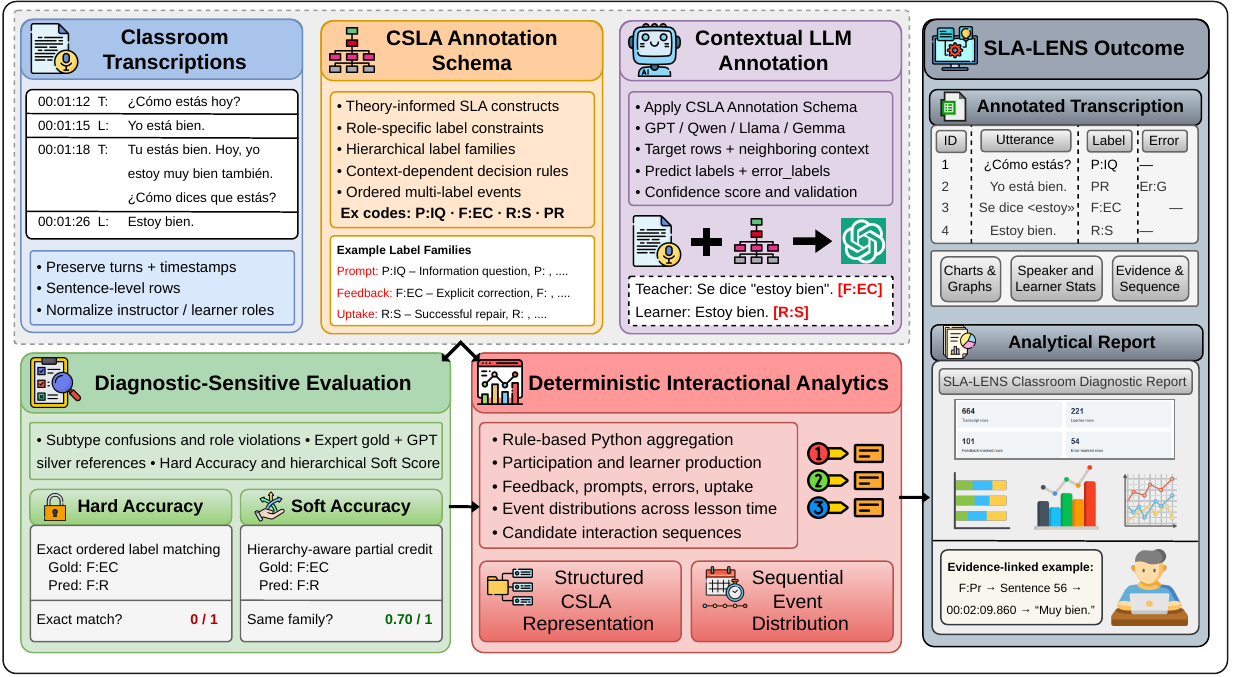}
    \caption{
    Overview of the SLA-LENS end-to-end framework.
    Classroom transcripts are structured and annotated using
    contextual LLMs guided by a theory-informed CSLA schema.
    Annotation quality is assessed through diagnostic-sensitive evaluation, while a deterministic analytics layer computes
    descriptive interactional measures and candidate event
    associations. SLA-LENS produces an annotated transcript
    and an evidence-linked analytical report with participant-level
    statistics, temporal visualizations, and supporting transcript
    examples.
    }
    \label{fig:csla_pipeline}
\end{figure*}

Although annotations are assigned to individual transcript rows, their interpretation often requires surrounding conversational context. Moreover, identical expressions can serve different functions: \textit{Muy bien}, for example, may indicate praise or regulate conversational flow. Authentic transcripts introduce further challenges through fragmented speech, code-switching, transcription noise, and imbalanced annotation categories. 
Large language models (LLMs) offer a promising approach to contextual annotation by conditioning structured predictions on multi-turn discourse. Open-weight models additionally support task-specific adaptation and institutionally controlled deployment, potentially reducing the cost, reproducibility, and privacy constraints associated with proprietary APIs \citep{liagkou2024,leon2026}. We investigate both proprietary and open-weight models, including task-specific fine-tuning, for SLA-informed classroom annotation. 

We introduce \textbf{SLA-LENS}, an end-to-end framework that transforms L2 classroom transcripts into structured annotations and evidence-linked descriptive analytics (Fig~\ref{fig:csla_pipeline}). 
These outputs help educators and researchers locate interactional patterns for closer inspection and can inform pedagogical reflection, curriculum planning, and program development. Extending these analyses with speaker metadata and temporal records could support comparisons of classroom participation and instructional experiences across time and social groups and further support the investigation of learning progression. Our main contributions are:
\begin{itemize}
    \setlength{\itemsep}{0pt}
    \setlength{\parskip}{0pt}
    \setlength{\parsep}{0pt}
    \setlength{\topsep}{2pt}

    \item \textbf{SLA-LENS, an end-to-end transcript-to-analytics framework} that combines contextual LLM annotation with deterministic, evidence-linked descriptive reporting.

    \item \textbf{A theory-informed SLA annotation schema and corpus} capturing classroom events in authentic online L2 Spanish interaction.

    \item \textbf{Open-weight model adaptation and evaluation}, including QLoRA fine-tuning and a diagnostic-sensitive evaluation framework.

    \item \textbf{A reproducible interactional reporting tool} providing evidence-based analysis for diverse research and educational applications.
\end{itemize}

\section{Related Work}

\paragraph{Interactionist SLA and the Computational Scalability Gap}

Interactionist SLA research has established the importance of negotiation of meaning, corrective feedback, modified output, and learner uptake in language development \citep{long1996,swain2005,gass2015,li2010}. Studies have examined how feedback strategies influence learner responses and acquisition \citep{ellis2006,sheen2010}. \citet{lyster1997} provides an influential framework linking learner production, instructor feedback, uptake, and repair, distinguishing strategies such as explicit correction, recasts, clarification requests, metalinguistic feedback, and elicitation. Analyzing these phenomena requires context-sensitive discourse coding. Conversation-analytic approaches provide detailed accounts of classroom interaction \citep{gardner2019classroom}, while quantitative observational studies rely on manually identifying and classifying relevant exchanges. Such events can be relatively sparse: across \citet{lyster1997} and \citet{panova2002}, approximately 39 feedback moves, 20 uptake instances, and nine repairs were identified per hour on average. Similarly, \citet{loewen2004} reported approximately 29 focus-on-form episodes per hour. These findings illustrate the substantial annotation effort required to examine interactional patterns across larger classroom corpora.

\paragraph{Computational and LLM-Based Approaches to SLA}

Computational research has supported learner corpora, grammatical-error analysis, automated assessment, and feedback generation \citep{mcenery2019corpus}, including identification of discourse features in L2 writing \citep{eguchi2024nlp}. Educational NLP has also addressed instructional discourse analysis \citep{demszky2023}, computational teacher-learner feedback interactions \citep{castrogarcia2025}, and adaptive language-learning support \citep{beaudin2007,turapova2025}. However, these approaches primarily examine written text, instructional support, or controlled interactions rather than jointly assessing multiple SLA phenomena in naturally occurring classroom conversations. Recent work has increasingly investigated LLMs as providers of corrective feedback, explanations, and conversations \citep{crusan2026,dong2026,pitychoutis2026,kamelabad2026}. SLA-LENS instead uses LLMs to analyze existing human interaction, identifying context-dependent classroom events for subsequent quantitative analysis. This introduces challenges involving diarization, transcription noise, rare labels, and overlapping communicative functions.

\paragraph{Evidence-Linked Descriptive Reporting}

Learning analytics research highlights the importance of interpretable educational visualizations tailored to different users and analytical goals \citep{schwendimann2017,valle2021,kaliisa2023}. Building on this perspective, SLA-LENS connects computational discourse annotation with reproducible descriptive reporting, including event distributions, temporal patterns, participant-level comparisons, and supporting transcript evidence. Separating annotation from deterministic aggregation allows the same structured data to support pedagogical reflection, corpus-based SLA research, and broader exploratory analyses across sessions or participant groups. These descriptive measures provide a foundation for future longitudinal investigations, metadata integration, and inferential statistical analysis.

\section{SLA-LENS: Transcript-to-Analytics Framework}
\label{sec:framework}

\begin{table}[t]
\centering
\small
\setlength{\tabcolsep}{3pt}
\begin{tabular}{lr}
\hline
\textbf{Corpus statistic} & \textbf{Count} \\
\hline
Transcripts & 40 \\
Sentence rows & 21,835 \\
Approximate words & 106,336 \\
Instructor rows & 14,648 \\
Learner rows & 7,187 \\
Single-label rows & 21,106 \\
Multi-label rows & 716 \\
Feedback-positive rows & 4,487 \\
Uptake/repair-positive rows & 721 \\
Learner-error rows & 2,078 \\
Observed classroom labels & 26 \\
\hline
\end{tabular}
\caption{Overview of the SLA-LENS annotated corpus. Counts are derived from GPT-generated annotations rather than independently validated gold labels.}
\label{tab:corpus_stats}
\end{table}

SLA-LENS is an end-to-end framework that transforms L2 classroom transcripts into structured SLA annotations and evidence-linked descriptive interactional analytics. As illustrated in Fig~\ref{fig:csla_pipeline}, the framework comprises three stages: transcript structuring, contextual LLM annotation, and deterministic aggregation and reporting. It produces two complementary outputs: an inspectable annotated transcript and a standalone analytical report supporting descriptive analysis of classroom interaction.

\subsection{Classroom Data and Transcript Representation}
\label{sec:data}

\begin{table*}[t]
\centering
\small
\renewcommand{\arraystretch}{1.15}
\begin{tabular}{p{2.2cm} p{2.6cm} p{1.0cm} p{8.5cm}}
\hline
\textbf{Family} & \textbf{Labels} & \textbf{Role} & \textbf{Function and example} \\
\hline

Prompt
& P:I, P:IQ, P:CQ
& Instructor
& Elicits learner production through instructions, information-seeking questions, or clarification questions.
\newline\emph{Example:} ``So, the first question for today is: ¿cuál es el tema de la Unidad 7?'' [Eng: ``What is the topic of Unit 7?''] \\

Prompt response
& PR
& Learner
& Responds to the preceding prompt.
\newline\emph{Example:} ``El tema de la Unidad 7 es la justicia.'' [Eng:``The topic of Unit 7 is justice.''] \\

Feedback
& F:EC, F:R, F:CR, F:EL, F:MF, F:CC, F:Pr, F:Ack, F:NE
& Instructor
& Responds to learner production through correction, reformulation, elicitation, metalinguistic info, evaluation, or acknowledgment.
\newline\emph{Example following a lexical error:} ``Justicia means justice.'' \\

Other instructor functions
& A, RN, CM, CoM
& Instructor
& Answers learner questions, reacts to content, manages classroom logistics, or regulates conversational flow.
\newline\emph{Example after a learner describes a large family:} ``That is a big family!'' \\

Uptake / repair outcome
& R:S, R:IR, R:Ack, R:NR
& Learner
& Characterizes a learner response following feedback as successful repair, incomplete repair, acknowledgment, or no uptake.
\newline\emph{Example following correction:} ``Mi hermana tiene trece años.''[Eng: ``My sister is thirteen years old.''] \\

Student prompt
& SP, SP:I, SP:IQ, SP:CQ
& Learner
& Initiates or extends interaction through questions, instructions, or requests for clarification.
\newline\emph{Example:} ``¿Cómo se dice \textit{kite} en español?''[Eng:``How do you say \textit{kite} in Spanish?''] \\

Other learner functions
& CM, CoM
& Learner
& Manages classroom logistics or conversational flow.
\newline\emph{Example:} ``I'm trying to look at your email, but I don't see it.'' \\

\hline
Learner error
& Er:G, Er:L, Er:M, Er:EU
& Learner
& Marks grammatical, lexical, multiple, or contextually inappropriate English production.
\newline\emph{Example:} ``Mi hermana \textbf{tengo} trece años.''
\newline\emph{Intended:} ``My sister is thirteen years old.'' \\

\hline
\end{tabular}
\caption{Condensed CSLA annotation schema used by SLA-LENS. Classroom-event and learner-error annotations are stored independently. Full definitions, decision rules, and examples are provided in Appendix~\ref{app:guidelines}.}
\label{tab:csla_schema}
\end{table*}

Our corpus comprises 40 online L2 Spanish classroom transcripts from two instructional settings: structured classroom instruction and language-tutoring sessions. Classroom sessions are comparatively instructor-led, with explicit learning tasks and instructional feedback, whereas tutoring sessions emphasize conversational practice and more open-ended learner production. The transcripts originate from online recordings and automatically generated transcriptions, which were minimally corrected before annotation to address transcription artifacts while preserving non-target learner production, code-switching, hesitations, and incomplete utterances. Each transcript is represented as an ordered sequence of sentence-level rows containing identifiers, timestamps, speaker identity, speaker role, and transcript text. Speaker roles are normalized as \textit{instructor} or \textit{learner}, while surrounding rows provide conversational context for annotation. The corpus contains 21,835 sentence rows and approximately 106K words, with instructor and learner speech accounting for 67.1\% and 32.9\%, respectively (Table~\ref{tab:corpus_stats}). The GPT-based annotation pipeline assigns a single classroom-event label to most rows, while 3.3\% contain multiple labels and 9.5\% receive learner-error annotations. Label frequencies are imbalanced: P: and PR occur frequently, whereas several feedback and uptake subtypes are rare. This motivates the label-sensitive evaluation described in Section~\ref{sec:critical_results}.

\subsection{CSLA Annotation Schema}
\label{sec:schema}

The CSLA annotation schema operationalizes SLA-relevant interactional phenomena into structured labels suitable for human and LLM annotation. Developed with SLA expertise, it was iteratively refined through analysis of authentic classroom transcripts and annotation disagreements. Rather than exhaustively describing classroom discourse, the schema captures observable pedagogical and interactional events that support reproducible descriptive analysis. Table~\ref{tab:csla_schema} summarizes the label families, with detailed definitions, decision rules, and examples provided in Appendix~\ref{app:guidelines}.

\paragraph{Two annotation dimensions.}
The schema separates \textit{classroom events} from \textit{learner errors}. The \texttt{labels} dimension captures communicative functions such as prompts, feedback, learner responses, and classroom management, while the independent \texttt{error\_labels} dimension identifies observable learner-language errors. Thus, a learner response may simultaneously receive \texttt{PR} (learner Prompt Response) and \texttt{Er:G} (Grammatical Error). Error annotations are restricted to learner production and exclude apparent errors attributable to transcription artifacts.

\paragraph{Context and communicative function.}
Labels represent an utterance's interactional function rather than its surface form. For example, \textit{repite} (`repeat') may constitute an instructional prompt when initiating a task or corrective feedback when responding to erroneous learner production. Similarly, learner production following an instructional prompt receives \texttt{PR}, whereas responses within feedback sequences belong to the \texttt{R} family. These distinctions require local conversational context even though annotations are assigned at the sentence-row level.

\paragraph{Priority and role constraints.}
Explicit priority rules resolve competing interpretations: Feedback takes precedence over other instructor functions when responding to learner production, while Uptake/Repair takes precedence over other learner functions within feedback sequences. Thus, $F >$ other instructor functions and $R >$ other learner functions. Speaker-role constraints further restrict label assignments, preventing instructor-specific labels on learner rows and vice versa. These rules guide annotation and subsequent evaluation.

\paragraph{Hierarchy and multi-label events.}
Several labels belong to hierarchical families, such as \texttt{F:R} (Recast) and \texttt{F:EC} (Explicit Correction) within Feedback. This structure supports evaluation that distinguishes within-family confusion from errors across interactional functions. Multiple labels are assigned only when a row contains independently identifiable events, with label order reflecting their occurrence. For example, \textit{Muy bien, ahora inténtalo otra vez} receives \texttt{F:Pr, F:EL}. Multi-label annotations account for 3.3\% of corpus rows.

\paragraph{Learner errors.}
The error taxonomy distinguishes single grammatical (\texttt{Er:G}), single lexical (\texttt{Er:L}), and multiple errors (\texttt{Er:M}), alongside substantially English production when Spanish is contextually expected (\texttt{Er:EU}). English use is not automatically erroneous, particularly when permitted by the instructional context or used for classroom management (e.g., \textit{We can't hear you, Professor}). Punctuation, formatting, and identifiable ASR artifacts are excluded.

\subsection{End-to-End Analysis Pipeline}
\label{sec:pipeline}

SLA-LENS transforms input transcripts into the structured representation described in Section~\ref{sec:data}, preserving sentence-level timestamps, speaker identities, and turn information while normalizing speaker roles as \textit{instructor} or \textit{learner}. The annotation engine processes transcript rows with surrounding conversational context and the CSLA annotation guidelines. For each target row, the LLM predicts classroom-event \texttt{labels} and learner \texttt{error\_labels}, alongside confidence and a brief rationale. Structured-output validation checks label inventories, speaker-role constraints, and sentence alignment before annotations are merged with the original transcript. GPT-4.1 serves as the reference annotation backend, while open-weight alternatives are evaluated in Section~\ref{sec:evaluation_data}. Implementation and prompting details are provided in Appendix~\ref{app:pipeline}.

The annotated transcript serves both as an independent research artifact and as input to SLA-LENS's deterministic reporting layer. This layer computes descriptive interactional measures, including participant contributions, prompt and feedback distributions, learner-error patterns, uptake/repair outcomes, and their variation across lesson time. Explicit proximity-based rules additionally identify candidate associations between prompts and responses, learner errors and feedback, and feedback and subsequent uptake. All measures are linked to supporting sentence identifiers, timestamps, and transcript excerpts, allowing users to inspect the underlying evidence. SLA-LENS produces two complementary outputs: an exportable annotated CSV and a standalone analytical report containing visualizations, participant-level summaries, and an annotation glossary. Table~\ref{tab:diagnostic_measures} summarizes the supported measures, while Section~\ref{sec:diagnostic_output} describes the reporting interface.

\begin{table}[t]
\centering
\small
\setlength{\tabcolsep}{3pt}
\renewcommand{\arraystretch}{1.08}
\begin{tabularx}{\columnwidth}{@{}p{2.05cm} X@{}}
\hline
\textbf{Analysis dimension} & \textbf{Descriptive measures} \\
\hline
Participation &
Instructor and learner contributions; individual learner production. \\

Learner errors &
Frequency, type distribution, and learner-normalized error incidence. \\

Pedagogical events &
Prompt and feedback subtype distributions; uptake/repair outcomes. \\

Temporal patterns &
Distribution of annotated events across lesson intervals. \\

Error $\rightarrow$ Feedback &
Candidate associations between error-marked learner production and subsequent feedback. \\

Feedback $\rightarrow$ Uptake &
Candidate associations between instructor feedback and subsequent learner uptake/repair. \\

Prompt $\rightarrow$ Response &
Candidate associations between instructor prompts and learner responses. \\

Evidence &
Supporting sentence IDs, timestamps, speakers, and transcript excerpts. \\
\hline
\end{tabularx}
\caption{Descriptive interactional measures supported by SLA-LENS. Sequence-based measures represent proximity-derived candidate associations rather than verified pedagogical episodes.}
\label{tab:diagnostic_measures}
\end{table}

\section{Annotation Engine Evaluation}
\label{sec:annotation_eval}

The reliability of SLA-LENS's descriptive analytics depends on the quality of its underlying annotations. We therefore evaluate the contextual annotation engine independently of the deterministic reporting layer, examining both annotation agreement with expert judgments and the feasibility of replacing the proprietary backend with locally deployable open-weight models.

\subsection{Gold and Silver Evaluation Sets}
\label{sec:evaluation_data}

We evaluate annotation quality using two complementary
reference sets drawn from five transcripts held out from
model adaptation (2,621 sentence-level rows).
The \textbf{gold set} consists of one of these transcripts
(653 rows), whose classroom event and learner error
annotations were manually reviewed and adjudicated by
three expert annotators. It serves as the primary reference
for assessing agreement with human judgments. The \textbf{silver set} comprises GPT-generated
annotations for all five held-out transcripts. It provides
a larger-scale measure of how closely open-weight models
reproduce the reference annotation engine on unseen
classroom interactions. Gold and silver results are
reported separately, as agreement with automatically
generated silver labels does not establish agreement
with expert annotations.

\subsection{Models and Experimental Setup}
\label{sec:models}

We compare GPT-4.1, the reference annotation backend, against three open-weight models: Qwen3-8B, Llama-3.1-8B, and Gemma 4 12B. All models perform the same contextual annotation task, predicting classroom-event \texttt{labels} and learner \texttt{error\_labels} from transcript rows with speaker-role information and neighboring conversational context. The annotation guidelines, label inventory, and output format are kept consistent across systems, with model-specific formatting adjustments where necessary.

\paragraph{Open-Weight Model Adaptation.}
We additionally fine-tune Qwen3-8B using Quantized Low-Rank Adaptation (QLoRA) on the GPT-annotated training corpus. The approach combines 4-bit quantization with trainable low-rank adapters, enabling task-specific adaptation without updating all pretrained parameters. Training examples preserve the structured transcript representation, conversational context, and annotation targets. The resulting model is evaluated against both expert gold annotations and held-out GPT-generated silver annotations. The detailed training hyperparameters and inference settings are provided in Appendix~\ref{app:training_details}.


\subsection{Evaluation Metrics}
\label{sec:evaluation}

The CSLA schema is hierarchical, speaker-role-dependent, and occasionally multi-label. Strict exact-match accuracy therefore treats errors of different severity equally. We complement exact agreement with a hierarchical similarity score and targeted detection metrics for interactional phenomena central to SLA-LENS reporting.

\paragraph{Hard Accuracy.}
Hard Accuracy measures the proportion of rows whose predicted classroom-event labels exactly match the reference annotations. For multi-label rows, both label identity and ordering must agree.

\paragraph{Hierarchical Soft Score.}
We compute a row-level Soft Score that assigns partial credit according to the CSLA label hierarchy (Table ~\ref{tab:soft_metric}) Exact matches receive full credit, within-family subtype errors receive partial credit, and predictions from another family compatible with the speaker role receive lower credit. Role-incompatible predictions receive no credit. Multi-label cases additionally account for partially recovered labels, unsupported predictions, and incorrect label ordering. The final score is averaged across rows and scaled to 0-100.

\begin{table}[t]
\centering
\small
\setlength{\tabcolsep}{4pt}
\renewcommand{\arraystretch}{1.08}
\begin{tabular}{lr}
\hline
\textbf{Prediction relation} & \textbf{Credit} \\
\hline
Exact label & 1.00 \\
Correct family, wrong subtype & 0.70 \\
Correct role, wrong family & 0.25 \\
Role-incompatible prediction & 0.00 \\
Correct labels, incorrect order & $\times 0.85$ \\
\hline
\end{tabular}
\caption{Base credits for hierarchical annotation similarity. Multi-label scoring additionally accounts for partial matches and unsupported labels.}
\label{tab:soft_metric}
\end{table}

\label{sec:results}
\begin{table*}[t]
\centering
\small
\setlength{\tabcolsep}{5pt}
\renewcommand{\arraystretch}{1.08}
\begin{tabular}{lrrrrrr}
\hline
\textbf{Model} &
\textbf{Hard Acc.} &
\textbf{Soft Score} &
\textbf{Feedback F1} &
\textbf{Uptake F1} &
\textbf{Error F1} &
\textbf{Error Hard Acc.} \\
\hline
GPT-4.1
& 76.72 & 84.12 & 87.57 & 76.64 & 81.40 & 96.63 \\
Gemma 4 12B
& 54.30 & 67.20 & 65.10 & 45.80 & 61.70 & 90.80 \\
Llama-3.1-8B
& 58.70 & 70.90 & 69.80 & 50.60 & 65.40 & 91.90 \\
Qwen3-8B
& 62.40 & 73.80 & 73.50 & 55.20 & 69.10 & 92.80 \\
Qwen3-8B (QLoRA)
& 69.80 & 79.60 & 81.20 & 65.40 & 75.80 & 94.90 \\
\hline
\end{tabular}
\caption{Annotation performance against the expert-reviewed gold transcript. All values are percentages except Soft Score, which is on a 0-100 scale.}
\label{tab:gold_results}
\end{table*}

\begin{table*}[t]
\centering
\small
\setlength{\tabcolsep}{5pt}
\renewcommand{\arraystretch}{1.08}
\begin{tabular}{lrrrrrr}
\hline
\textbf{Model} &
\textbf{Hard Acc.} &
\textbf{Soft Score} &
\textbf{Feedback F1} &
\textbf{Uptake F1} &
\textbf{Error F1} &
\textbf{Error Hard Acc.} \\
\hline
Gemma 4 12B
& 66.80 & 76.40 & 76.20 & 58.70 & 70.30 & 94.20 \\
Llama-3.1-8B
& 70.20 & 79.10 & 79.80 & 63.40 & 74.10 & 95.30 \\
Qwen3-8B
& 74.10 & 82.20 & 83.60 & 67.90 & 77.80 & 96.10 \\
Qwen3-8B (QLoRA)
& 81.30 & 87.10 & 89.20 & 76.50 & 84.10 & 97.20 \\
\hline
\end{tabular}
\caption{Agreement with GPT-generated annotations on five held-out silver transcripts. Silver agreement measures reproduction of GPT annotations rather than agreement with expert judgments.}
\label{tab:silver_results}
\end{table*}

\paragraph{Diagnostic-Sensitive Evaluation.}
We separately report detection F1 for three interactional phenomena important to downstream analysis. \textbf{Feedback F1} measures detection of any \texttt{F:*} subtype, while \textbf{Uptake/Repair F1} evaluates the learner-response family (\texttt{R:S}, \texttt{R:IR}, \texttt{R:Ack}, \texttt{R:NR}). \textbf{Error Detection F1} measures whether a learner row contains any error annotation, independently of its subtype. We additionally report \textbf{Error Hard Accuracy}, requiring exact agreement on the \texttt{error\_labels} field.  Row-level scores can also support subtype-specific analyses. Further details of these metrics are provided in Appendix ~\ref{app:evaluator}.

\section{Annotation Results}
\label{sec:mainresults}

We evaluate the SLA-LENS annotation engine against an expert-reviewed gold transcript and five held-out silver transcripts using the metrics in Section~\ref{sec:evaluation}. Gold evaluation measures agreement with human annotations, while silver evaluation measures agreement with GPT-generated labels. We compare GPT-4.1 with Qwen3-8B, Llama-3.1-8B, and Gemma 4 12B, alongside Qwen3-8B adapted using quantized low-rank adaptation (QLoRA).

\subsection{Overall Performance}
\label{sec:overall_results}

Tables~\ref{tab:gold_results} and~\ref{tab:silver_results} summarize annotation performance against expert and silver references, respectively. On the gold set, GPT-4.1 achieves 76.72\% Hard Accuracy and a hierarchical Soft Score of 84.12. Feedback detection reaches an F1 of 87.57\%, while uptake/repair detection is lower at 76.64\%. Error detection achieves 81.40\% F1, compared with 96.63\% Error Hard Accuracy, reflecting the prevalence of error-free rows. The open-weight comparison examines whether smaller, locally deployable models can recover the CSLA annotation structure and whether task-specific adaptation improves their performance.
On the gold set, QLoRA adaptation improves
Qwen3-8B Hard Accuracy from 62.40\% to
69.80\% (+7.40 percentage points) and
Feedback F1 from 73.50\% to 81.20\%
(+7.70 points). The adapted model also
shows higher agreement with the GPT-generated
silver references.

\subsection{Diagnostic-Critical Label Performance}
\label{sec:critical_results}

Overall accuracy can obscure performance on less frequent but analytically important events. We therefore separately evaluate Feedback, Uptake/Repair, and learner-error detection, which directly influence SLA-LENS's descriptive interactional measures. Family-level F1 assesses event identification. GPT-4.1 identifies Feedback more reliably than Uptake/Repair, suggesting that learner-response annotation remains a particularly challenging component. Such differences matter downstream: missed feedback can distort feedback distributions, while missed uptake can underrepresent learner responses following correction. 

\subsection{Annotation Error Analysis}
\label{sec:annotation_error_analysis}

Qualitative inspection of model--expert disagreements reveals challenges involving contextual interpretation, transcription artifacts, and annotation ambiguity. In one example, GPT classified \textit{muy bien} (`very good') as praise, whereas all three adjudicators classified it as a conversational flow marker (CoM): the teacher was introducing the first instructional question rather than evaluating learner production. All three adjudicators also assigned Unknown to \textit{Sounds Avocado}, identifying it as a transcription artifact, while GPT assigned a discourse category. Other cases reflect uncertainty in the reference interpretation. Two adjudicators classified \textit{s\'i} (`yes') as CoM, while the third interpreted it as the teacher's response to an implicit learner prompt. These distinctions matter for downstream reporting: misclassifying a flow marker as praise inflates praise counts, while classifying a transcription artifact introduces an unsupported event. Adjudicator disagreement also shows that some utterances permit multiple plausible interpretations. Together, these cases support reporting complementary evaluation measures and retaining transcript evidence for human inspection.



\begin{quote}
\footnotesize
\setlength{\parskip}{2pt}

\noindent\textbf{Teacher:}
\textit{Before we start, do you have any questions or issues---anything I can help you with?}

\noindent\textbf{Learner:}
\textit{No, no.}

\noindent\textbf{Teacher:}
\textit{Okay, perfect.}

\noindent\textbf{Teacher:}
\textit{Bien.} (`Good.')

\noindent\textbf{Teacher:}
\textit{Entonces.} (`So.')

\noindent\textbf{Teacher:}
\textit{Muy bien, la primera pregunta del d\'ia de hoy.}\\
{\scriptsize English: All right, the first question of the day.}
\end{quote}



\begin{quote}
\footnotesize
\setlength{\parskip}{2pt}

\noindent\textbf{Learner:}
\textit{So I don't know how \ldots{} I don't know how to say it.}

\noindent\textbf{Learner:}
\textit{I'm sorry.}

\noindent\textbf{Teacher:}
\textit{Vamos!} (`Come on!')

\noindent\textbf{Learner:}
\textit{Like \ldots}

\noindent\textbf{Teacher:}
\textit{S\'i.} (`Yes.')
\end{quote}

\noindent These analyses highlight why annotation quality cannot be summarized by exact-match accuracy alone. For SLA-LENS, the significance of a prediction error also depends on which interactional measure it affects. This motivates reporting hierarchical agreement alongside targeted detection scores and retaining the transcript-level evidence.

\section{Interactional Analytics and Reporting}
\label{sec:diagnostic_output}

Beyond structured annotation, SLA-LENS generates a standalone analytical report for researchers, educators, and other practitioners studying language-learning interaction. Constructed deterministically from the annotated transcript without additional LLM-generated interpretation, the report provides reproducible descriptive summaries in HTML and PDF formats. 

\subsection{Interactional Measures}

SLA-LENS organizes its analyses into three dimensions: (i) participation, including instructor-learner contributions and individual learner production; (ii) pedagogical events, including prompts, feedback strategies, learner errors, and uptake/repair outcomes; and (iii) interactional sequences, identifying candidate associations between learner errors, instructor feedback, and subsequent responses. The report presents frequency distributions, learner-level comparisons, normalized error incidence, and temporal visualizations. Sequence associations follow explicit conversational proximity rules and are not treated as verified pedagogical exchanges.

\subsection{Evidence-Grounded Analytical Report}

The standalone report presents session-level
summary cards showing transcript size, instructor and
learner contributions, and counts of feedback, errors,
and uptake, alongside participant information and
data-quality indicators. Learner-level visualizations
compare participation, approximate word production,
normalized error incidence, and error-category
distributions across speakers. Additional views display
pedagogical-event frequencies, lesson timelines, and
candidate associations between errors, feedback, and
learner responses.
To support interpretation, the report includes a
compact CSLA label glossary and an evidence index
linking annotations to sentence IDs, timestamps,
speakers, and transcript excerpts. These descriptive outputs highlight patterns
for investigation rather than establishing learner
proficiency or instructional effectiveness.
Appendix~\ref{app:diagnostic_report} presents screenshots
of the report interface and learner-level analyses;
the complete report is available in the supplementary
repository.

\section{Conclusion}

We introduced \textbf{SLA-LENS}, an end-to-end framework that moves beyond transcription to make second-language classroom interaction computationally observable and interpretable. Combining a theory-informed annotation schema, contextual LLM annotation, and deterministic reporting, SLA-LENS transforms classroom transcripts into structured, evidence-linked descriptive analytics. Our evaluation investigates the reliability of proprietary and open-weight annotation models, including task-specific adaptation, while the reporting layer demonstrates how learner errors, feedback, uptake, participation, and temporal patterns can be explored at scale. Beyond supporting pedagogical reflection, SLA-LENS provides a reproducible foundation for corpus-based SLA research and computational humanities. Future work will extend the framework toward longitudinal and cross-classroom comparisons, integration of learner proficiency and sociolinguistic metadata, and statistical modeling of interactional patterns and language development. By connecting automated discourse annotation with transparent analytical reporting, SLA-LENS offers a step toward more accessible, scalable, and empirically grounded analysis of language-learning interaction. The resulting evidence can help teachers reflect on their pedagogical practices and inform curriculum designers' decisions about classroom activities and instructional support.

\section*{Acknowledgments}

We thank Megan Michelle Smith for her valuable contributions
to the development of this work and for her assistance
during the research process. Her input and support
helped inform aspects of the study.

\section*{Limitations}

SLA-LENS is currently developed and evaluated on English--Spanish instructional interaction, limiting its demonstrated generalizability to other language pairs, proficiency levels, and educational settings. The corpus does not systematically incorporate learner proficiency, demographic or sociolinguistic metadata, or longitudinal learner trajectories, preventing controlled comparisons across learner groups and direct investigation of language development. Furthermore, evaluation relies on a small subset of expert-reviewed gold transcript and LLM-generated silver annotations, which may introduce annotation bias and limit the reliability of broader performance estimates. The framework operates on sentence-level transcripts and does not explicitly identify complete interactional episodes or source-target relationships among learner errors, feedback, and repair. Sequence-based associations therefore rely on proximity heuristics rather than verified discourse links. Transcription and speaker-attribution errors may also propagate into annotation and reporting. Finally, the current analytics are descriptive and do not establish proficiency, learning gains, causal effects, or instructional effectiveness; the practical utility of the reports has not yet been validated through systematic user studies.

\section*{Ethical Considerations}

SLA-LENS analyzes classroom interactions involving human participants and therefore requires appropriate consent, privacy safeguards, and responsible data handling. The study received institutional IRB approval, and all
participating instructors and learners provided informed
consent for recording and research use. Protecting participant anonymity is a priority throughout the workflow, including transcription, annotation, data analysis, aggregation, and reporting. LLM-generated annotations may misclassify non-standard learner language, code-switching, or transcription artifacts as errors, potentially introducing biases into participant-level summaries. Accordingly, SLA-LENS provides descriptive, evidence-linked analyses rather than automated judgments of learner proficiency or instructional effectiveness, and findings should be interpreted with human oversight. AI-assisted writing tools (e.g., ChatGPT and Grammarly) were used solely for grammatical refinement and clarity of presentation.

\bibliography{custom}

\clearpage

\appendix

\appendix

\section{CSLA Annotation Guidelines}
\label{app:guidelines}

This appendix summarizes the operational guidelines used to produce CSLA annotations in SLA-LENS. The schema contains two independent annotation dimensions: \texttt{labels}, representing classroom interaction events, and \texttt{error\_labels}, representing observable learner-language errors. The full annotation manual provides additional examples and edge cases.

\subsection{General Principles}

Annotations represent the \textit{communicative function} of an utterance in conversational context rather than its grammatical form. Only phenomena supported by the available transcript are annotated. A single classroom-event label is preferred unless a row contains multiple independently observable events, in which case labels follow their temporal order.

Two priority rules resolve common ambiguities:

\noindent
\textbf{Instructor:} $F >$ other instructor functions. \quad
\textbf{Learner:} $R >$ other learner functions.

Thus, instructor utterances responding to learner production are classified as Feedback when appropriate, even when their surface form resembles a prompt. Likewise, learner responses within feedback sequences are classified as Uptake/Repair rather than ordinary prompt responses.

\paragraph{Unknown Function.}
\texttt{UNK} is assigned when an utterance's communicative function cannot be determined reliably from the available context. This includes severely fragmented or unintelligible transcript segments and transcription artifacts that do not support a defensible interactional interpretation. \texttt{UNK} is preferred over speculative assignment of another classroom-event label.

\subsection{Instructor Functions}

Instructor functions include prompts, feedback, answers, reactions, classroom management, and conversation management.

\paragraph{Prompts.}
A Prompt initiates or elicits learner production rather than responding to a preceding learner error.

\begin{table}[t]
\centering
\footnotesize
\setlength{\tabcolsep}{3pt}
\renewcommand{\arraystretch}{1.10}
\begin{tabularx}{\columnwidth}{@{}p{0.85cm} X@{}}
\hline
\textbf{Label} & \textbf{Definition} \\
\hline
P:I  & Instruction or task intended to elicit learner production. \\
P:IQ & Information-seeking question requesting new information. \\
P:CQ & Clarification question outside a corrective-feedback sequence. \\
\hline
\end{tabularx}
\caption{Instructor prompt subtypes.}
\label{tab:prompt_types}
\end{table}

\noindent\textbf{Example.}
\textit{Describe el personaje.} $\rightarrow$ \texttt{P:I}.

\paragraph{Feedback.}
Feedback responds to learner production or the learner's language-learning process. Whenever Feedback is assigned, a specific subtype is required.

\begin{table}[t]
\centering
\footnotesize
\setlength{\tabcolsep}{3pt}
\renewcommand{\arraystretch}{1.10}
\begin{tabularx}{\columnwidth}{@{}p{0.85cm} X@{}}
\hline
\textbf{Label} & \textbf{Definition} \\
\hline
F:EC  & Explicit correction that overtly provides the target form. \\
F:R   & Recast that reformulates learner production without explicitly identifying the error. \\
F:CR  & Clarification request following problematic learner production. \\
F:EL  & Elicitation prompting the learner to supply or complete the corrected form. \\
F:MF  & Metalinguistic information about the linguistic problem without directly supplying the correction. \\
F:CC  & Confirmation check of a proposed interpretation or corrected form. \\
F:Pr  & Positive evaluation of learner language or performance. \\
F:Ack & Acknowledgment or repetition of a correct learner form without explicit praise. \\
F:NE  & Negative evaluation without supplying the correction. \\
\hline
\end{tabularx}
\caption{Instructor feedback subtypes.}
\label{tab:feedback_types}
\end{table}

Feedback takes precedence over Prompt when the instructor utterance responds to learner production.

\noindent\textbf{Example.}
Following a learner error,
\textit{Di ``tiene''.} $\rightarrow$ \texttt{F:EC},
not \texttt{P:I}.

\paragraph{Other Instructor Functions.}

\begin{table}[t]
\centering
\footnotesize
\setlength{\tabcolsep}{3pt}
\renewcommand{\arraystretch}{1.10}
\begin{tabularx}{\columnwidth}{@{}p{0.85cm} X@{}}
\hline
\textbf{Label} & \textbf{Definition} \\
\hline
A   & Answers a learner question without functioning as corrective feedback. \\
RN  & Reacts to the meaning, content, or emotional aspect of learner production without evaluating linguistic accuracy. \\
CM  & Manages lesson organization, technology, participation, or logistics. \\
CoM & Regulates conversational flow through minimal expressions such as \textit{okay}, \textit{mhm}, or \textit{vale}. \\
\hline
\end{tabularx}
\caption{Other instructor functions.}
\label{tab:other_instructor}
\end{table}

\subsection{Learner Functions}

Learner functions distinguish responses to instructional prompts from responses occurring within corrective-feedback sequences.

\paragraph{Prompt Response.}
\texttt{PR} marks learner production in response to an instructor prompt, question, or assigned task. The response may be correct, incorrect, complete, or incomplete. Learner errors are annotated independently.

\noindent\textbf{Example.}
T: \textit{¿Cuál es tu comida favorita?}
S: \textit{La pizza.} $\rightarrow$ \texttt{PR}.

\paragraph{Student Prompt.}
The \texttt{SP} family captures learner-initiated prompts, questions, and clarification requests.

\begin{table}[t]
\centering
\footnotesize
\setlength{\tabcolsep}{3pt}
\renewcommand{\arraystretch}{1.10}
\begin{tabularx}{\columnwidth}{@{}p{0.85cm} X@{}}
\hline
\textbf{Label} & \textbf{Definition} \\
\hline
SP    & General learner-initiated prompt when a specific subtype cannot be determined reliably. \\
SP:I  & Directs another participant to perform an action. \\
SP:IQ & Requests new information. \\
SP:CQ & Requests clarification, repetition, explanation, or lexical assistance. \\
\hline
\end{tabularx}
\caption{Learner-initiated prompt categories.}
\label{tab:student_prompt_types}
\end{table}

\paragraph{Response and Uptake.}
The \texttt{R} family characterizes learner responses within feedback sequences. Without preceding instructor feedback, learner production is not classified as an \texttt{R} event.

\begin{table}[t]
\centering
\footnotesize
\setlength{\tabcolsep}{3pt}
\renewcommand{\arraystretch}{1.10}
\begin{tabularx}{\columnwidth}{@{}p{0.85cm} X@{}}
\hline
\textbf{Label} & \textbf{Definition} \\
\hline
R:S   & Successful repair that resolves the problem targeted by feedback. \\
R:IR  & Attempted repair that does not fully resolve the targeted problem. \\
R:Ack & Acknowledges feedback without producing the corrected form. \\
R:NR  & No learner uptake before the interaction moves on. \\
\hline
\end{tabularx}
\caption{Learner uptake and repair outcomes.}
\label{tab:uptake_types}
\end{table}

The distinction between \texttt{PR} and \texttt{R} depends on the preceding instructor event. Responses to instructional prompts receive \texttt{PR}, whereas learner responses within feedback sequences receive an \texttt{R} subtype.

\noindent\textbf{Example.}
T: \textit{No. Di ``tiene''.} $\rightarrow$ \texttt{F:EC};
S: \textit{Tiene.} $\rightarrow$ \texttt{R:S}.

\noindent\textbf{Annotation boundary.}
The \texttt{R:NR} category represents the absence of uptake following feedback. Because SLA-LENS assigns labels to existing transcript rows rather than generating separate episode-level annotations, this category requires particular care. Its assignment must be supported by the available interactional evidence and the row-level annotation convention.

\paragraph{Interaction Management.}
Learner \texttt{CM} concerns lesson logistics, technology, or participation, while \texttt{CoM} marks minimal expressions that regulate conversational flow without contributing new instructional content.

\subsection{Continuation and Multi-label Rules}

Details, elaborations, and continuations inherit the communicative function of the event they continue rather than receiving a separate explanation label.

\noindent\textbf{Continuation example.}
T: \textit{Provide at least two Spanish names.}
$\rightarrow$ \texttt{P:I};
T: \textit{By Spanish names I mean names such as Juan or María.}
$\rightarrow$ \texttt{P:I}.

Multiple labels are assigned only when a row contains more than one independently observable classroom event. Their order follows the order of occurrence.

\noindent\textbf{Multi-label example.}
\textit{Muy bien, ahora inténtalo otra vez.}
$\rightarrow$ \texttt{F:Pr, F:EL}.

Secondary labels are not added merely because an utterance permits multiple interpretations. When the temporal order or independence of proposed functions cannot be established reliably, annotators should avoid speculative additional labels.

\subsection{Learner Error Annotation}

Learner errors are annotated independently of classroom-event functions and only when supported by observable learner production. Errors attributable to ASR, punctuation, capitalization, or formatting are excluded.

\begin{table}[t]
\centering
\footnotesize
\setlength{\tabcolsep}{3pt}
\renewcommand{\arraystretch}{1.10}
\begin{tabularx}{\columnwidth}{@{}p{0.85cm} X@{}}
\hline
\textbf{Label} & \textbf{Definition} \\
\hline
Er:G  & One grammatical error involving morphology, syntax, agreement, word order, or another grammatical feature. \\
Er:L  & One incorrect, missing, malformed, or non-Spanish lexical item in otherwise Spanish production. \\
Er:M  & More than one observable grammatical or lexical error. \\
Er:EU & Substantially English production when Spanish is contextually expected. \\
\hline
\end{tabularx}
\caption{Learner-language error categories.}
\label{tab:error_types}
\end{table}

\begin{table*}[t]
\centering
\footnotesize
\setlength{\tabcolsep}{4pt}
\renewcommand{\arraystretch}{1.12}
\begin{tabularx}{\textwidth}{@{}p{2.2cm} X X@{}}
\hline
\textbf{Distinction} &
\textbf{Decision rule} &
\textbf{Example} \\
\hline

Prompt vs.\ Feedback &
Use \texttt{P} when the instructor initiates new learner production. Use \texttt{F} when responding to, evaluating, or facilitating correction of learner production. &
\textit{Di una oración.} $\rightarrow$ \texttt{P:I};
after an error, \textit{Di ``tiene''.} $\rightarrow$ \texttt{F:EC}. \\

PR vs.\ R &
Use \texttt{PR} after an instructional prompt and an \texttt{R} subtype within a feedback sequence. &
T: \textit{¿Cómo estás?}
S: \textit{Bien.} $\rightarrow$ \texttt{PR};
T: \textit{Di ``tiene''.}
S: \textit{Tiene.} $\rightarrow$ \texttt{R:S}. \\

F:Pr vs.\ RN &
Use \texttt{F:Pr} for positive evaluation of learner performance and \texttt{RN} for reactions to the content or situation described. &
S: \textit{Estoy cansado.}
T: \textit{Lo siento.} $\rightarrow$ \texttt{RN}. \\

F:Ack vs.\ CoM &
Use \texttt{F:Ack} when acknowledging learner production as a pedagogical response. Use \texttt{CoM} when the expression primarily regulates conversational flow. &
T: \textit{Muy bien.} $\rightarrow$ \texttt{F:Ack} or \texttt{CoM}, depending on preceding learner production and context. \\

CM vs.\ CoM &
Use \texttt{CM} for lesson logistics, technology, participation, or classroom organization. Use \texttt{CoM} for minimal discourse-management expressions. &
\textit{Voy a compartir mi pantalla.}
$\rightarrow$ \texttt{CM};
\textit{Okay.} $\rightarrow$ \texttt{CoM}. \\

Er:L vs.\ Er:EU &
Use \texttt{Er:L} for an isolated non-Spanish lexical item and \texttt{Er:EU} for substantially English production when Spanish is expected. &
\textit{Mi mamá lives en California.}
$\rightarrow$ \texttt{Er:L};
\textit{My mother lives in California.}
$\rightarrow$ \texttt{Er:EU}. \\

\hline
\end{tabularx}
\caption{Key contextual distinctions in the CSLA annotation guidelines.}
\label{tab:ambiguity_rules}
\end{table*}

\paragraph{Error Decision Hierarchy.}
Error categories are assigned using the following ordered procedure:

\begin{enumerate}[leftmargin=*, nosep]
    \item If Spanish is expected and the utterance is substantially English, assign \texttt{Er:EU}.
    \item Otherwise, if more than one observable error occurs, assign \texttt{Er:M}.
    \item If exactly one grammatical error occurs, assign \texttt{Er:G}.
    \item If exactly one lexical error occurs, assign \texttt{Er:L}.
    \item Otherwise, leave \texttt{error\_labels} empty.
\end{enumerate}

English is not automatically considered erroneous. English used for classroom management, explicitly permitted by the instructor, or required by the task receives no \texttt{Er:EU}. An isolated English lexical item embedded in otherwise Spanish production is generally \texttt{Er:L}.

When an apparent error cannot be reliably distinguished from an ASR, segmentation, or speaker-attribution artifact, annotation remains conservative. Error labels describe observable evidence in the available transcript rather than independently verified learner proficiency.

\noindent\textbf{Examples.}
\textit{Mi hermana tengo trece años.}
$\rightarrow$ \texttt{Er:G};
\textit{Yo borrow un libro.}
$\rightarrow$ \texttt{Er:L};
\textit{Los casa es grande.}
$\rightarrow$ \texttt{Er:M};
\textit{The boy is running.}
$\rightarrow$ \texttt{Er:EU}, when Spanish is expected.

\subsection{Key Ambiguity Rules}

Several annotation distinctions depend on conversational context rather than surface form. Table~\ref{tab:ambiguity_rules} summarizes the principal decision rules.

\section{SLA-LENS Annotation Pipeline Details}
\label{app:pipeline}

\subsection{Transcript Processing}

The annotation pipeline accepts \texttt{.vtt},
\texttt{.txt}, and \texttt{.docx} transcript files.
Timestamped cues are parsed into speaker blocks
and segmented into sentence-level rows using
sentence-final punctuation and Spanish inverted
punctuation. Each row retains its sentence ID,
turn number, source turn number, timestamps,
speaker identity, and utterance text. Sentences
derived from the same source cue retain its
timestamps. The annotation engine processes 12 target
rows per batch, accompanied by up to five
preceding and two following rows as
conversational context. Only target rows
receive annotations. The prompt combines
the CSLA annotation guidelines, role-specific
constraints, priority rules, and structured
output requirements. The current LLM implementations use temperature
0, \texttt{top\_p}=1, and a maximum of
4,000 completion tokens. The complete
prompts and JSON schema are provided in
the supplementary repository.

\begin{tcolorbox}[
    colback=gray!7,
    colframe=gray!55,
    boxrule=0.4pt,
    arc=1mm,
    left=5pt,
    right=5pt,
    top=5pt,
    bottom=5pt,
    breakable
]
\footnotesize

\textbf{Prompt for Contextual CSLA Annotation}

\medskip
\textbf{System Prompt (excerpt).}

You are an expert linguistic annotator for
classroom interaction transcripts.

Follow the CSLA annotation guidelines.
Annotate only the TARGET ROWS and use
CONTEXT ROWS to interpret the surrounding
interaction. Treat speaker roles as
authoritative, do not invent missing content,
and do not silently correct transcription
errors. Return only valid JSON.

\medskip
\textbf{User Prompt (excerpt).}

\textbf{Input:} Context rows and target rows
containing sentence IDs, timestamps, speaker
roles, and utterance text.

\textbf{Goal:} Assign SLA-relevant annotations
to each target row using the CSLA guidelines.

\textbf{Key Rules:}
\begin{itemize}[leftmargin=*, nosep]
    \item Identify communicative functions
    using conversational context.
    \item Prioritize instructor Feedback
    when responding to learner production.
    \item Prioritize learner Uptake/Repair
    within feedback sequences.
    \item Apply speaker-role constraints
    and ordered multi-label rules.
    \item Annotate observable learner errors
    independently of classroom events.
    \item Use \texttt{UNK} when evidence
    is insufficient.
\end{itemize}

\textbf{Output:} One structured JSON object
containing annotations for all target rows,
including label assignments, confidence,
and brief rationales.

\medskip
The complete prompt, label inventory,
decision rules, and output schema are
provided in the supplementary repository.
\end{tcolorbox}

\subsection{Structured Output and Validation}

The model returns a JSON object containing
one annotation per target row. Each annotation
includes \texttt{sentence\_id},
\texttt{turn\_number}, \texttt{speaker},
\texttt{sentence}, \texttt{labels},
\texttt{error\_labels},
\texttt{confidence}, and \texttt{rationale}. A strict JSON schema restricts allowable
label values and required fields. Returned
batches are checked for the expected number
and ordered IDs of target rows.

\section{Model and Training Details}
\label{app:training_details}

This appendix provides additional implementation details
for the proprietary and open-weight annotation models
evaluated in Section~\ref{sec:models}. All models perform
the same CSLA annotation task, predicting classroom-event
\texttt{labels} and learner \texttt{error\_labels}.

\subsection{Model Checkpoints and Inference}

We evaluate GPT-4.1 as the reference annotation system
alongside three open-weight models: Qwen3-8B,
Llama-3.1-8B-Instruct, and Gemma-4-12B-it.
GPT-4.1 uses the \texttt{gpt-4.1-2025-04-14}
checkpoint, while the open-weight models are based
on the instruction-tuned checkpoints
\texttt{Qwen/Qwen3-8B},
\texttt{meta-llama/Llama-3.1-8B-Instruct}, and
\texttt{google/gemma-4-12B-it}, respectively.
To maintain a consistent annotation setup, all models
receive the same CSLA guidelines, speaker-role
information, and conversational context, consisting
of 12 target rows, five preceding rows, and two
following rows. Inference uses temperature $0$,
\texttt{top\_p}=1, and a maximum of 4000 generated
tokens. All models are instructed to produce structured
annotations containing classroom-event labels,
learner-error labels, confidence, and brief rationales.
The complete prompts and inference configurations
are provided in the supplementary repository.

\subsection{Training Data Construction}

The corpus contains 40 GPT-annotated classroom
transcripts comprising 21,835 sentence-level rows.
Of these, 35 transcripts (19,214 rows) are designated
for open-weight model adaptation, while five
transcripts (2,621 rows) are held out for silver
evaluation, out of which one was used to construct the gold transcript. The training partition contains 12,920
instructor rows and 6,294 learner rows, with 19,202
rows containing at least one classroom-event label.
The silver partition contains 1,728 instructor rows
and 893 learner rows, of which 2,620 have classroom
annotations. The expert-reviewed gold transcript
contains 653 sentence rows and provides an
independent reference for human-annotation agreement.
Training examples are constructed from GPT-generated
CSLA annotations and include target utterances,
speaker roles, conversational context, and structured
annotation outputs. The evaluation
transcripts are reserved from model adaptation to
support evaluation on unseen classroom interaction.

\subsection{QLoRA Fine-Tuning}

We adapted Qwen3-8B using Quantized
Low-Rank Adaptation (QLoRA) on the GPT-annotated
CSLA training corpus. The base model is loaded
using 4-bit NF4 quantization with bfloat16
computation. Low-rank adapters are applied to
the attention and feed-forward projection
modules (\texttt{q\_proj}, \texttt{k\_proj},
\texttt{v\_proj}, \texttt{o\_proj},
\texttt{gate\_proj}, \texttt{up\_proj},
and \texttt{down\_proj}), with rank $r=16$,
scaling parameter $\alpha=32$, and dropout 0.05. Fine-tuning uses a learning rate of $2\times10^{-4}$,
the paged 8-bit AdamW optimizer, and a cosine
learning-rate schedule with a 3\% warmup ratio.
It is trained for three epochs with a
per-device batch size of 2 and gradient
accumulation of 8, yielding an effective batch
size of 16 on one GPU. The maximum sequence
length is 8,192 tokens. Training is done on a single NVIDIA A100
80 GB GPU using Hugging Face Transformers,
PEFT, TRL, and bitsandbytes.

\subsection{Evaluation and Reproducibility}

All prompt-only and adapted models are evaluated
on the same expert-reviewed gold and held-out silver
transcripts. Predictions are aligned by sentence ID
and evaluated using Hard Accuracy, hierarchical Soft
Score, and diagnostic-sensitive detection metrics
as defined in Section~\ref{sec:evaluation}.
The gold and silver results are reported separately
because silver annotations originate from the GPT-based
reference system rather than an independent human review.
The supplementary repository includes the annotation
and evaluation scripts.

\section{Hierarchical Evaluation Implementation}
\label{app:evaluator}

This appendix specifies the evaluation procedure
used for classroom-event and learner-error
annotations. Gold and predicted rows are aligned
using \texttt{sentence\_id} when available, with
row-position alignment used otherwise. Classroom-event
labels are evaluated as ordered sequences, while
learner-error annotations are scored independently.

\subsection{Exact-Match Evaluation}

For each row $i$, let $G_i$ and $P_i$ denote the
ordered gold and predicted classroom-event label
sequences. Hard Accuracy is defined as

\begin{equation}
\mathrm{HardAcc} =
\frac{1}{N}\sum_{i=1}^{N}
\mathbf{1}[G_i=P_i].
\end{equation}

where $N$ is the number of aligned rows.
Both label identity and order must match.
Error Hard Accuracy is calculated analogously
using the ordered \texttt{error\_labels} sequences.

\subsection{Hierarchical Soft Score}

The Soft Score assigns partial credit according
to label-family membership and speaker-role
compatibility. For a gold label $g$, predicted
label $p$, and row role $r$, pairwise credit is

\begin{equation}
c(g,p,r)=
\begin{cases}
1.00 & g=p,\\
0.70 & \mathrm{fam}(g)=\mathrm{fam}(p),\\
0.25 & \mathrm{role}(g,r)\land
        \mathrm{role}(p,r),\\
0.00 & \text{otherwise}.
\end{cases}
\end{equation}

Here, $\mathrm{fam}(\cdot)$ identifies the CSLA
label family, and $\mathrm{role}(l,r)$ indicates
whether label $l$ is compatible with the row's
speaker role. The family condition applies to
distinct valid labels within the same family.
Role-neutral labels (\texttt{CM}, \texttt{CoM},
and \texttt{UNK}) are compatible with both
instructor and learner roles.
For multi-label rows, the evaluator selects
the maximum-credit one-to-one matching
$M_i$ between gold and predicted labels.
Unmatched labels receive zero credit, and
the matching score is normalized by the
larger label-set cardinality:

\begin{equation}
A_i =
\frac{
\max_{M_i}
\sum_{(g,p)\in M_i}c(g,p,r_i)
}{
\max(|G_i|,|P_i|)
}.
\end{equation}

When both sequences are empty, $A_i=1$;
when exactly one is empty, $A_i=0$.
This formulation penalizes both missing
and unsupported labels.
An ordering penalty is applied only when
at least two exactly shared labels occur
in a different relative order between
the gold and predicted sequences:

\begin{equation}
S_i =
100\,A_i
\begin{cases}
0.85 & \text{if shared-label order differs},\\
1.00 & \text{otherwise}.
\end{cases}
\end{equation}

The final Soft Score is the mean row-level
score:

\begin{equation}
\mathrm{SoftScore} =
\frac{1}{N}\sum_{i=1}^{N}S_i.
\end{equation}

\subsection{Worked Scoring Examples}

Table~\ref{tab:soft_examples} illustrates
the scoring behavior for instructor rows.

\begin{table}[t]
\centering
\footnotesize
\setlength{\tabcolsep}{3pt}
\renewcommand{\arraystretch}{1.10}
\begin{tabularx}{\columnwidth}{@{}X X r@{}}
\hline
\textbf{Gold} & \textbf{Prediction} &
\textbf{Score} \\
\hline
\texttt{F:EC} & \texttt{F:EC} & 100.0 \\
\texttt{F:EC} & \texttt{F:R} & 70.0 \\
\texttt{F:EC} & \texttt{P:I} & 25.0 \\
\texttt{F:EC} & \texttt{PR} & 0.0 \\
\texttt{F:Pr, P:IQ} & \texttt{F:Pr} & 50.0 \\
\texttt{F:Pr, P:IQ} & \texttt{P:IQ, F:Pr}
& 85.0 \\
\hline
\end{tabularx}
\caption{Illustrative hierarchical Soft Scores
for instructor-role rows. Scores reflect
label-family credit, role compatibility,
multi-label coverage, and ordering penalties.}
\label{tab:soft_examples}
\end{table}

\subsection{Diagnostic-Sensitive Detection}

Feedback, Uptake/Repair, and learner-error
detection are evaluated as separate row-level
binary classification tasks. A row is positive
when its annotation contains at least one
label from the corresponding family:
\texttt{F:*} for Feedback,
\texttt{R:*} for Uptake/Repair, and
\texttt{Er:*} for learner errors.
For each category, precision, recall, and
F1 are calculated from row-level true
positives, false positives, and false
negatives:

\begin{equation}
P=\frac{TP}{TP+FP}, \qquad
R=\frac{TP}{TP+FN},
\end{equation}

\begin{equation}
F_1=\frac{2PR}{P+R}.
\end{equation}

A prediction is considered correct at the
family-detection level regardless of subtype.
For example, predicting \texttt{F:R} instead
of \texttt{F:EC} counts as a true positive
for Feedback detection but receives only
partial credit under the hierarchical
Soft Score. Invalid classroom-event and error
labels are recorded separately, along with
speaker-role violations.

\begin{figure*}[t]
    \centering
    \includegraphics[
        width=0.96\textwidth,
        keepaspectratio
    ]{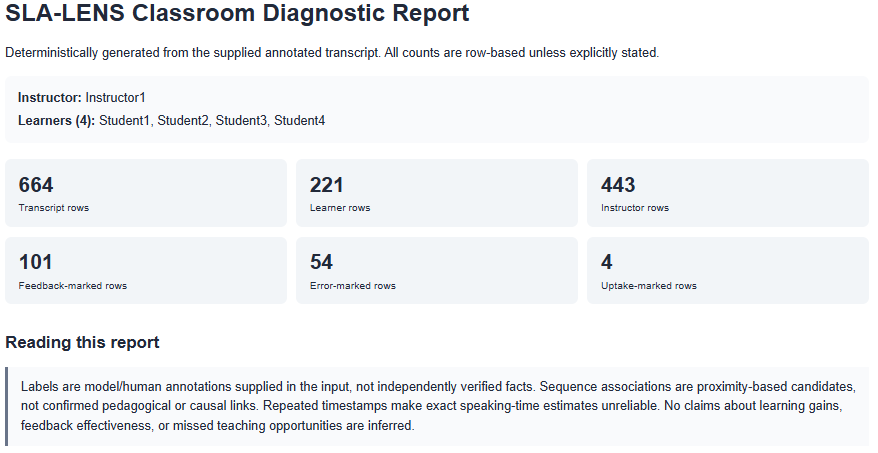}
    \caption{
    SLA-LENS analytical report overview showing anonymized
    participant information, session-level annotation counts,
    and guidelines for interpreting the descriptive outputs.
    All statistics are derived from the annotated transcript
    without additional LLM inference.
    }
    \label{fig:report_overview}
\end{figure*}

\begin{figure}[t]
    \centering
    \includegraphics[
        width=\columnwidth,
        keepaspectratio
    ]{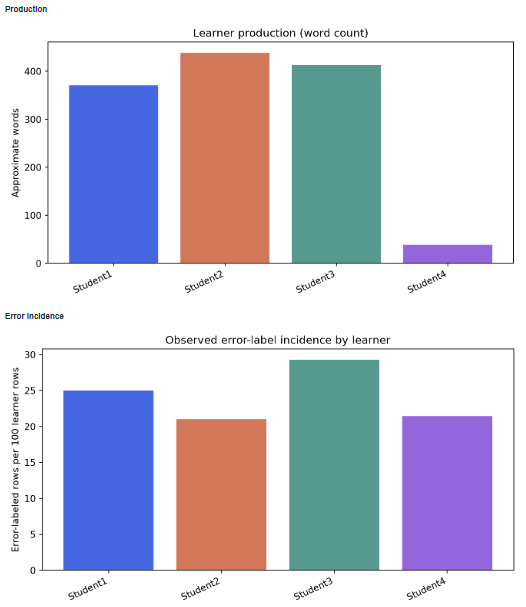}
    \caption{
    Learner-level descriptive analytics showing approximate
    word production (top) and error-labeled rows per 100
    learner rows (bottom). These measures account for
    differences in participation but do not independently
    indicate learner proficiency or language development.
    }
    \label{fig:learner_diagnostics}
\end{figure}

\section{SLA-LENS Analytical Report}
\label{app:diagnostic_report}

SLA-LENS generates standalone HTML and PDF reports from
annotated classroom transcripts using deterministic
Python-based aggregation. The report includes session-level
statistics, instructor and learner participation,
pedagogical-event distributions, learner-specific error
patterns, temporal visualizations, and candidate interactional
associations. An annotation glossary and evidence index
support independent interpretation and inspection of the
underlying transcript. Figure~\ref{fig:report_overview} shows the report interface
with session-level statistics and interpretation guidelines.
Figure~\ref{fig:learner_diagnostics} illustrates learner-level
comparisons of language production and annotated error
incidence.

\paragraph{Candidate Interaction Associations.}
SLA-LENS identifies candidate event associations using
deterministic row-proximity rules. For each source event,
the reporting algorithm searches forward for the first
eligible target event within four intervening transcript
rows, stopping if another source event occurs first.
Three association types are considered: learner error
to instructor corrective feedback, instructor feedback
to learner uptake, and instructor prompt to learner
prompt response. These associations are descriptive
proximity heuristics and do not establish verified
interactional episodes or causal relationships.

\paragraph{Evidence Traceability.}
Each reported annotation and candidate interaction
association can be traced to its supporting transcript
rows through sentence identifiers, timestamps,
speaker information, and original utterance text.
The generated evidence index allows users to inspect
these annotations in their conversational context.
Examples and the complete report are provided in
the supplementary repository.


\end{document}